\documentclass[10pt,twocolumn,letterpaper]{article}

\usepackage{cvpr}              \usepackage{bbding}

\definecolor{cvprblue}{rgb}{0.21,0.49,0.74}
\usepackage[pagebackref,breaklinks,colorlinks,citecolor=cvprblue]
{hyperref}

\def\paperID{} \def\confName{CVPR}
\def\confYear{2026}

\title{LEMON: A Large Endoscopic MONocular Dataset and Foundation Model for Perception in Surgical Settings}

\author{Chengan Che, Chao Wang, Tom Vercauteren, Sophia Tsoka, Luis C. Garcia-Peraza-Herrera \\
Department of Informatics, King's College London, UK \\
{\tt\small \{chengan.che, chao.wang, tom.vercauteren, sophia.tsoka, luis\_c.garcia\_peraza\_herrera\}@kcl.ac.uk}
}

\usepackage{booktabs}
\usepackage{graphicx}
\usepackage{amsmath}
\usepackage{multirow} \usepackage{tabularx}  \usepackage{listings}
\usepackage{hyperref}
\usepackage{subcaption}
\usepackage{pifont} 

\usepackage{algorithm}
\usepackage{algpseudocode}

\definecolor{MyGreen}{HTML}{E2EFDA}      \definecolor{MyBlue}{HTML}{DDEBF7} 
\definecolor{videoblue}{HTML}{4E85D9} \definecolor{imageorange}{HTML}{E97133} \usepackage{placeins}

\begin{document}
\maketitle
\begin{abstract}
Traditional open-access datasets focusing on surgical procedures are often limited by their small size, typically consisting of fewer than 100 videos and less than 30 hours of footage, which leads to poor model generalization. 
To address this data limitation, a new dataset called \textbf{LEMON} has been compiled using a novel aggregation pipeline that collects high-resolution videos from online sources.
%
% Featuring an extensive collection of over 4K surgical videos and more than 3 million high-quality images from multiple procedure types, LEMON offers a comprehensive resource surpassing existing alternatives in size and scope, including two novel tasks.
%
Featuring an extensive collection of over 4K surgical videos totaling 938 hours (85 million frames) of high-quality footage across multiple procedure types, LEMON offers a comprehensive resource surpassing existing alternatives in size and scope, including two novel downstream tasks.
To demonstrate the effectiveness of this diverse dataset, we introduce \textbf{LemonFM}, a foundation model pretrained on LEMON using a novel self-supervised augmented knowledge distillation approach.
%that achieves state-of-the-art results on four surgical downstream tasks and six different datasets.
%in surgical downstream tasks such as surgical phase recognition, action recognition, tool presence detection, and semantic segmentation.
%
%By combining self-supervised learning with a newly proposed augmented knowledge distillation method and a modern convolutional neural network architecture, 
LemonFM consistently outperforms existing surgical foundation models across four downstream tasks and six datasets, achieving significant gains in surgical phase recognition (+9.5pp, +9.4pp, and +8.4pp in Jaccard on AutoLaparo, M2CAI16, and Cholec80), surgical action recognition (+4.4pp in mAP on CholecT50), surgical tool presence detection (+5.3pp and +10.2pp in mAP on Cholec80 and GraSP), and surgical semantic segmentation (+10.3pp in mDice on CholecSeg8k).
%
%Moreover, LemonFM retains these advantages even when fine‑tuned with only half of the labeled data, demonstrating strong data efficiency.
%
LEMON and LemonFM will serve as foundational resources for the research community and industry, accelerating progress in developing autonomous robotic surgery systems and ultimately contributing to safer and more accessible surgical care worldwide.
Dataset, code, and models are publicly available at \href{https://github.com/visurg-ai/LEMON}{https://github.com/visurg-ai/LEMON}.

\end{abstract}

 \section{Introduction}
\label{sec:intro}

\begin{figure}[t!]
	\centering
	\includegraphics[width=\linewidth]{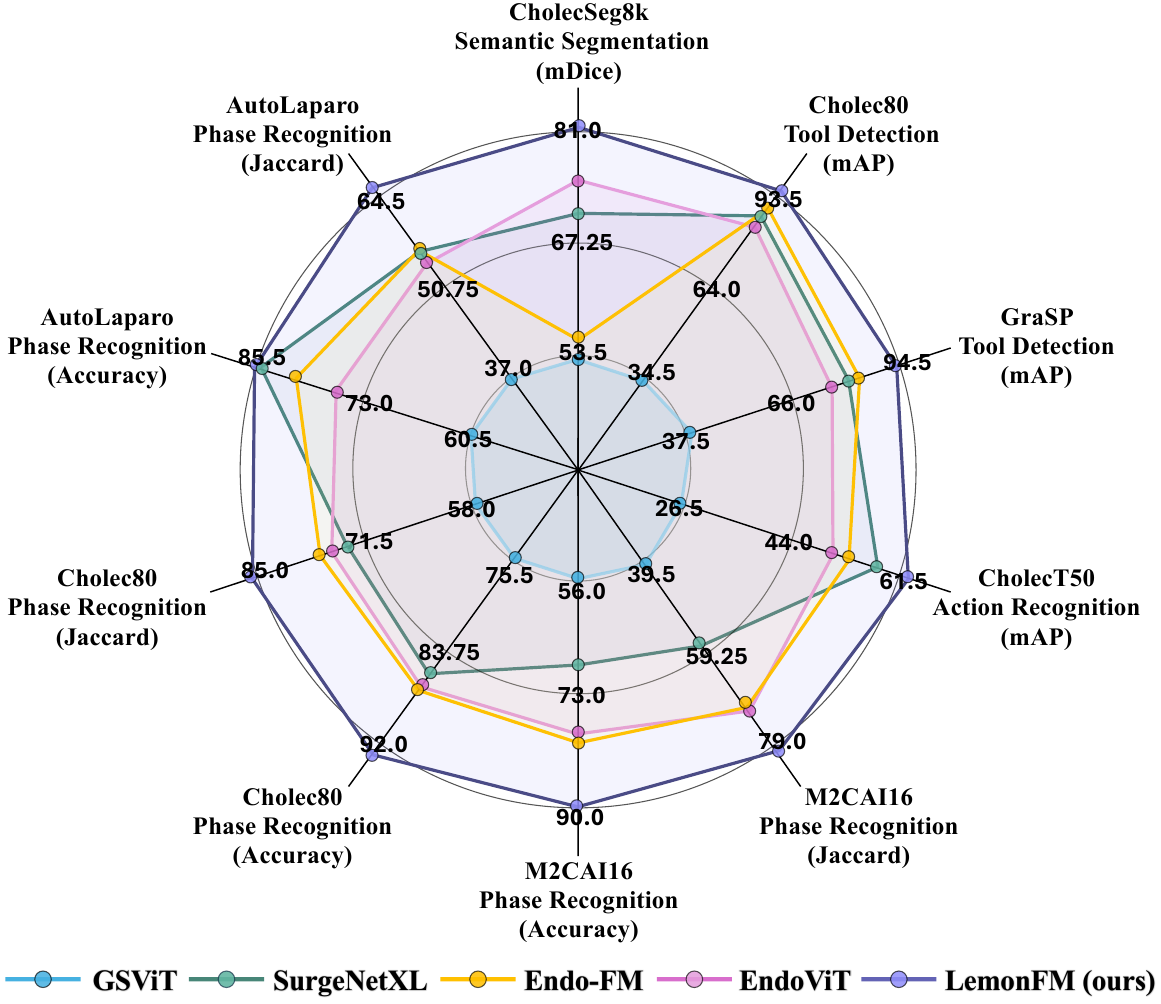}
	\caption{
        \textbf{Performance comparison between LemonFM, our foundation model pretrained on LEMON, and other surgical foundation models.} Data points are plotted relative to the numeric axis ticks.
        The full results are available in Table~\ref{tab:finetuned_classification} and~\ref{tab:segmentation}.
}
    \label{fig:result_downstreams}
\end{figure}

\begin{figure*}[t]
	\centering
	\includegraphics[width=.96\linewidth]{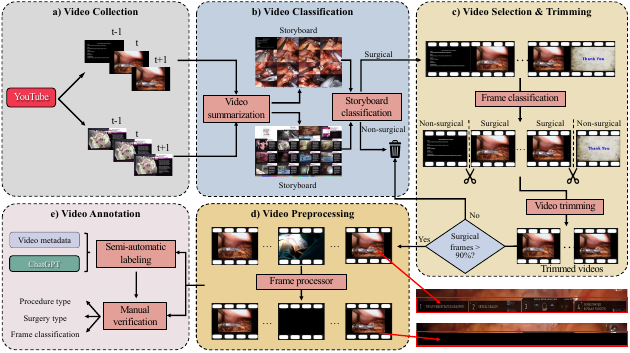}
	\caption{
        \textbf{Data curation pipeline.} 
(a) Surgical videos are collected from YouTube. 
(b) The videos are summarized as storyboards (trivial to annotate and classify). 
We train a storyboard classifier to identify videos with rich surgical content, and manually verify the selected videos. 
(c) During the video selection and trimming phase, a trained frame classifier categorizes each frame as either surgical or non-surgical, allowing us to trim non-surgical frame segments at the beginning and end of each video. 
We further filter videos by requiring at least 90\% of their frames to be surgical frames. 
(d) In the video preprocessing stage, we obliterate non-surgical frames and non-surgical regions within surgical frames.
(e) During the video annotation phase, we utilize the video title as a primary cue to determine the procedure and surgery type. 
In cases where the video titles do not explicitly include any of our procedure name keywords, we employ ChatGPT to match the titles with surgical procedure types. 
Both the final videos and their corresponding labels are manually quality controlled.
        }
	\label{fig:data_curation_pipeline}
\end{figure*}

Autonomous surgery has long been a promising area of research, with the potential to revolutionize the way surgical procedures are performed \cite{Hwang2023AutomatingHumans}.
It promises safer and more consistent operations, fewer complications, shorter operative times, lighter workloads for surgeons, and wider access to high‑quality care~\cite{Kim2025SRT-H:Learning}.
To achieve this goal, significant advancements in perception and computer vision are necessary, allowing robotic systems to accurately interpret and respond to the complexities of surgical environments~\cite{Maier-Hein2022SurgicalTranslation}.
However, training models for surgical perception is far from trivial due to the limited availability and high quality requirements of annotated data~\cite{Ward2021ChallengesAnnotation, Meireles2021SAGESVideo, NyangohTimoh2023AVideo}. 
The complexities of medical data and stringent regulations governing patient privacy create significant hurdles for collecting and annotating relevant information~\cite{Gilbert2021GeneratingSegmentation, Aouedi2023HandlingDirections}. 
The cumulative effect of all these challenges has resulted in significantly limited computer vision datasets for surgical applications, with many containing fewer than 10K images~\cite{Rueckert2024MethodsArt}. 
These datasets, which may suffice for validation on specific computer vision tasks and surgical procedures, fall short in providing an adequate foundation for training models that can generalize effectively across diverse surgical settings and datasets.

Recent advances in self-supervised learning have opened up new research avenues. Pretrained foundation models can significantly expedite progress, potentially reducing the need for annotated data by utilizing large-scale unlabeled datasets~\cite{Ramesh2023DissectingVision}.
Despite this progress, ongoing data collection challenges persist, motivating the exploration of two key aspects: the sufficiency of publicly available surgical videos as a foundation for building a comprehensive dataset, and the potential of self-supervised models trained on this dataset to achieve state-of-the-art performance in surgical downstream tasks. 
By addressing these aspects, we aim to provide valuable insights into the feasibility and efficacy of using publicly available online data for training machine learning models in the field of surgical vision analysis.

\begin{table}[t!]
\centering
\small
\setlength{\tabcolsep}{1.3pt}
\caption{
    \textbf{Comparison of LEMON's size to other existing public surgical datasets.} 
}
\label{tab:statistics_curated_videos}
    \begin{tabularx}{\linewidth}{@{}lccccc@{}}

    \hline
    \textbf{Dataset}    &  \textbf{Procedure} &  \textbf{Video} & \textbf{Hour} & \textbf{Frame} &  \textbf{Frame~(1\,fps)}  \\
    \hline
    M2CAI16~\cite{Stauder2016TheChallenge}  
    &  1  &  41 & $26$ & $2.4$M &  $95$K
    \\
    Cholec80~\cite{Twinanda2017EndoNet:Videos}                  
    &  1  &  80 & $51$ & $4.6$M &  $183$K
    \\
    CholecT50~\cite{Nwoye2022Rendezvous:Videos}
    &  1  &  50  & $28$ &$2.5$M & $101$K
    \\
    AutoLaparo~\cite{Wang2022AutoLaparo:Hysterectomy}
    &  1  &  21 & $23$ & $2.1$M & $83$K
    \\
    GraSP~\cite{Ayobi2024Pixel-WiseUnderstanding}
    &  $1$  &  $13$ & $33$ & $3.5$M &  $117$K
    \\
    GenSurgery~\cite{Schmidgall2024GeneralSurgery}
    &  $28$  &  $3182$ & $680$ & $70$M &  $2.4$M
    \\
    SurgeNetXL~\cite{Jaspers2026ScalingModels}     
    &  $23$  &  $3253$ & $576$ & $52$M &  $2.1$M
    \\
    \hline
    LEMON (\textbf{ours})   &  $\mathbf{35}$  &  $\mathbf{4194}$ &  $\mathbf{938}$ & $\mathbf{85}$\textbf{M} & $\mathbf{3.4}$\textbf{M}
    \\
    \hline
\end{tabularx}
\end{table}

\begin{figure*}[t!]
	\centering
	\includegraphics[width=.93\linewidth]{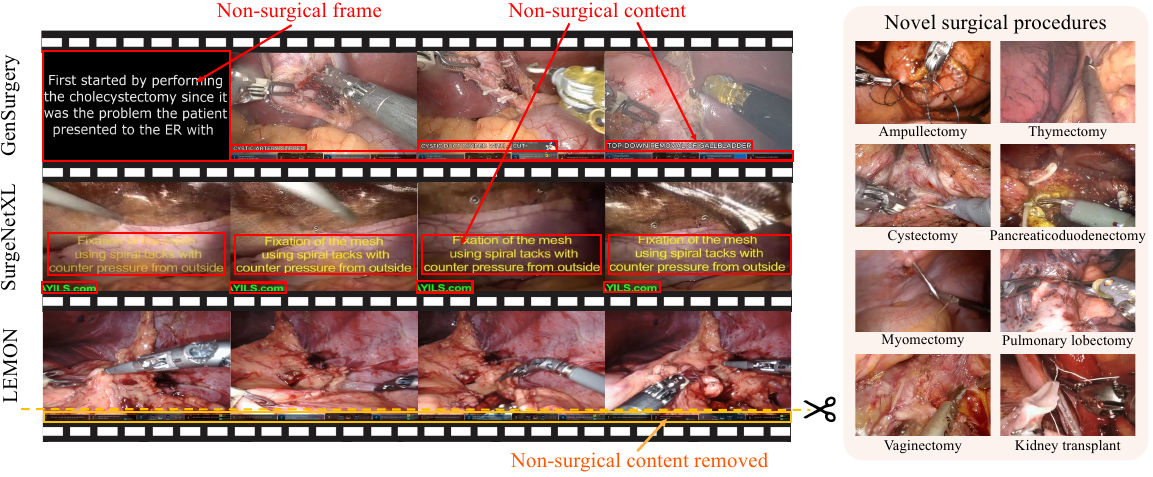}
	\caption{
        \textbf{Comparison between LEMON and other surgical datasets sourced from the web.}
        Video segments from each dataset illustrating the data curation differences (left).
Procedures which are not covered by both GenSurgery and SurgeNetXL or any other public dataset (right).
\textbf{Best viewed online.}
}
	\label{fig:lemon_comparison}
\end{figure*}

\noindent
\textbf{Contributions.}
To address the scarcity of high‑quality surgical data and the practical difficulty of collecting it, 
we propose a novel multi-stage data curation pipeline designed specifically for online surgical videos (Fig.~\ref{fig:data_curation_pipeline}).
In contrast to previous efforts that relied on limited, manually curated validation processes with datasets containing non-surgical content (Fig.~\ref{fig:lemon_comparison}), our pipeline employs a systematic approach to filter out non-surgical videos, remove non-surgical content from surgical videos, and verify the curated surgical videos and their annotations.
Building upon this data curation pipeline, we present LEMON, a large-scale dataset comprising more than 4K surgical videos totaling 938 hours (85 million frames) of high-quality footage (Table~\ref{tab:statistics_curated_videos}).
Additionally, we propose two novel supervised video classification tasks based on the accompanying annotations: multi-label surgical procedure classification and binary surgery type classification, and establish a leaderboard for them.
To the best of our knowledge, LEMON is the largest open access surgical dataset to date, both in total frames and videos and in the breadth of distinct surgical procedures it includes.
To model the features of surgical frames, we introduce LemonFM, a foundation model pretrained on LEMON that leverages self-supervised learning with our proposed augmented knowledge distillation method.
We evaluated LemonFM on six datasets: AutoLaparo~\cite{Wang2022AutoLaparo:Hysterectomy}, Cholec80~\cite{Twinanda2017EndoNet:Videos}, M2CAI16~\cite{Twinanda2017EndoNet:Videos}, GraSP~\cite{Ayobi2024Pixel-WiseUnderstanding}, CholecT50~\cite{Nwoye2022Rendezvous:Videos}, and CholecSeg8k~\cite{Hong2020CholecSeg8k:Cholec80}, covering four tasks including phase recognition, tool presence detection, action recognition, and semantic segmentation.
Under both linear probing and full fine-tuning protocols, LemonFM consistently surpasses existing surgical foundation models and task-specific architectures (Fig.~\ref{fig:result_downstreams}, Table~\ref{tab:linear_probing_classification}, \ref{tab:finetuned_classification}, \ref{tab:segmentation}).
Moreover, LemonFM retains these advantages even when fine‑tuned with only half of the labeled data.
These results showcase LemonFM’s strong generalization and data efficiency over diverse surgical scenarios and underscore the value of LEMON as a broadly useful resource for the community.
 \section{Related Work}
\label{sec:related_work}
\textbf{Surgical datasets.} 
Previous works have released open-access datasets aimed at addressing specific surgical vision tasks and procedures (Table \ref{tab:statistics_curated_videos}).
More recently, public surgical vision language resources have also appeared~\cite{Seenivasan2022Surgical-VQA:Transformer,Yuan2024AdvancingKnowledge, Li2024LLaVA-Surg:Learning}. 
Despite valuable, these datasets suffer from a narrow scope of surgical procedures and limited dataset sizes.
GenSurgery \cite{Schmidgall2024GeneralSurgery} and SurgeNetXL \cite{Jaspers2026ScalingModels} expand procedural coverage by compiling public videos, but the lack of a curation step to remove the non-surgical content from the public videos (Fig.~\ref{fig:lemon_comparison}) might introduce spurious features that can confound discriminative models.
Additionally, several private datasets exist, such as SVL~\cite{Yuan2024HecVL:Recognition} and SAGES CVS~\cite{Padoy2024SAGESChallenge}, but their restricted access limits their utility and broader adoption.
To overcome these limitations, we introduce a novel data curation pipeline that yields LEMON, the largest open-access surgical dataset to date, surpassing previous datasets in both the number of frames and videos and the diversity of distinct surgical procedures represented.

\noindent
\textbf{Self-supervised learning.}
Self-supervised learning (SSL)~\cite{Caron2021EmergingTransformers, Caron2020UnsupervisedAssignments, Chen2020ImprovedLearning, Su2025StreamlineDistillation, Chen2020ARepresentations, Du2026UnsupervisedDecoupling, Wang2023VideoMAEMasking, Sun2026HyperPoint:Space, Assran2025V-JEPAPlanning} has gained significant attention due to its ability to reduce reliance on large-scale annotated datasets, a benefit that is particularly relevant for surgical applications. Recently, Ramesh et al.~\cite{Ramesh2023DissectingVision} investigated the effectiveness of different SSL approaches for surgical applications. Their experimental results show that DINO~\cite{Caron2021EmergingTransformers} outperformed MoCov2~\cite{Chen2020ImprovedLearning}, SwAV~\cite{Caron2020UnsupervisedAssignments}, and SimCLR~\cite{Chen2020ARepresentations} in surgical applications.
As part of a broader effort to advance self-supervised surgical models, Hirsch et al.~\cite{Hirsch2023Self-supervisedAnalysis} built a private surgical dataset and trained a Masked Siamese Networks (MSN)~\cite{Assran2022MaskedLearning} on it.
Despite encouraging results, the work's wider applicability and reproducibility are limited by its reliance on private data and a relatively narrow evaluation scope: surgical phase recognition on Cholec80 and polyp characterization on PolypSet~\cite{Li2021ColonoscopyEvaluations}.
Similarly, in Endo-FM~\cite{Wang2023FoundationPre-train}, the authors proposed to train a student-teacher foundation model on a composite dataset that combined private and public data sources. 
Although showing promising results, the approach is also of limited accessibility. 
More recently, Batic et al.~\cite{Batic2024EndoViT:Images} created a dataset by merging nine public datasets. 
They trained a model based on masked autoencoders (MAE)~\cite{He2022MaskedLearners}, but showed limited performance, likely due to their reduced dataset size.
Self-supervised learning has also been explored in other surgical modalities~\cite{Fujii2024EgoSurgery-Phase:Videos, Basu2024FocusMAE:Autoencoders, Jamal2023SurgMAE:Analysis}, demonstrating its broad applicability across diverse surgical fields.

\section{Proposed LEMON Dataset}
\label{sec:LEMON}
To develop a robust surgical foundation model, it is essential to curate a large-scale, diverse dataset spanning the full spectrum of minimally invasive procedures. 
To address this, we propose LEMON, a comprehensive dataset comprising over 4K videos (938 hours) across 35 distinct procedure types, including both robotic-assisted and traditional non-robotic laparoscopies.

\subsection{Dataset Curation}
We summarize how we collected, curated, and annotated these videos below (Fig.~\ref{fig:data_curation_pipeline}), with comprehensive curation details, surgical background, and dataset validation provided in the supplementary material.

\noindent
\textbf{Video collection.} 
We first compiled a list of 35 distinct minimally invasive surgical procedure types~\cite{MayoKeyholeList2023}.
Inspired by works~\cite{Ephrat2018LookingSeparation,Carreira2017QuoDataset, Yang2019VideoSegmentation, Abu-El-Haija2016YouTube-8M:Benchmark, Monfort2020MomentsUnderstanding}, we collected our videos from YouTube. 
For each of the $35$ procedures, we searched for videos of their robotic and non-robotic variants using the strings ``robotic $<$procedure type$>$'' and ``laparoscopic $<$procedure type$>$'', respectively.
We run the video collection process until we had $500$ different videos collected for each procedure leading to a total of $\approx$18K raw videos.

\noindent
\textbf{Video classification.} 
The raw videos collected based on search keywords contained a diverse range of content, including patient testimonials and conference presentations. 
To avoid including non-surgical videos in our dataset, we classified the videos as surgical and non-surgical (Fig. ~\ref{fig:data_curation_pipeline}b). 
First, we created 4$\times$4-image video storyboards (i.e., a single image containing 16 key video frames) for each video using the method described in \cite{Garcia-Peraza-Herrera2023VideoSum:Summarization}. 
Second, a dataset of 2160 surgical and 1910 non-surgical video storyboards (Fig. \ref{fig:data_curation_pipeline}b)  was annotated.
Third, a binary video storyboard classifier (i.e., a ResNet18~\cite{He2016DeepRecognition}) was trained on this dataset and run over all the raw videos.
Finally, manual verification was performed on the videos classified as surgical.

\noindent
\textbf{Video selection and trimming.}
A video classified as surgical may still contain non-surgical frames (e.g., introductory and conclusion slides). Experimentally, we found that the start and end of the surgical footage can be reliably identified by finding the first and last three consecutive surgical frames (sampling the video at 1 fps).
Thus, we manually annotated a dataset of 7967 frames (5481 surgical and 2486 non-surgical), trained a surgical frame classifier (i.e., a ResNet18) on it, and used this model to: 1) trim non-surgical content present at the beginning and end of videos, and 2) discard videos containing more than 10\% of the frames estimated as non-surgical after trimming (Fig. \ref{fig:data_curation_pipeline}c).
\iffalse
These steps are detailed in the following paragraphs.
A ResNet18 was trained for the surgical/non-surgical video frame classification task.
To produce the annotations, the videos were sampled at one frame per second (fps), and we annotated 7967 frames, 5481 of which turned to be surgical, and 2486, non-surgical.
Most online videos contain introductory and conclusion slides. 
Experimentally, we found that the start and end of the surgical footage can be reliably identified by finding the first and last three consecutive frames classified as surgical by our surgical frame classifier (sampling the video at 1fps). 
Therefore, we discarded the non-surgical parts at the beginning and end of the collected videos (Fig.~\ref{fig:data_curation_pipeline}c). 
To retain only those videos that contain substantial surgical content, we rejected those videos that, when trimmed, still had more than 10\% of their frames classified by our frame classification model as non-surgical.
The remaining videos were further manually verified.
\fi

\begin{figure}[t!]
	\centering
	\includegraphics[width=\linewidth]{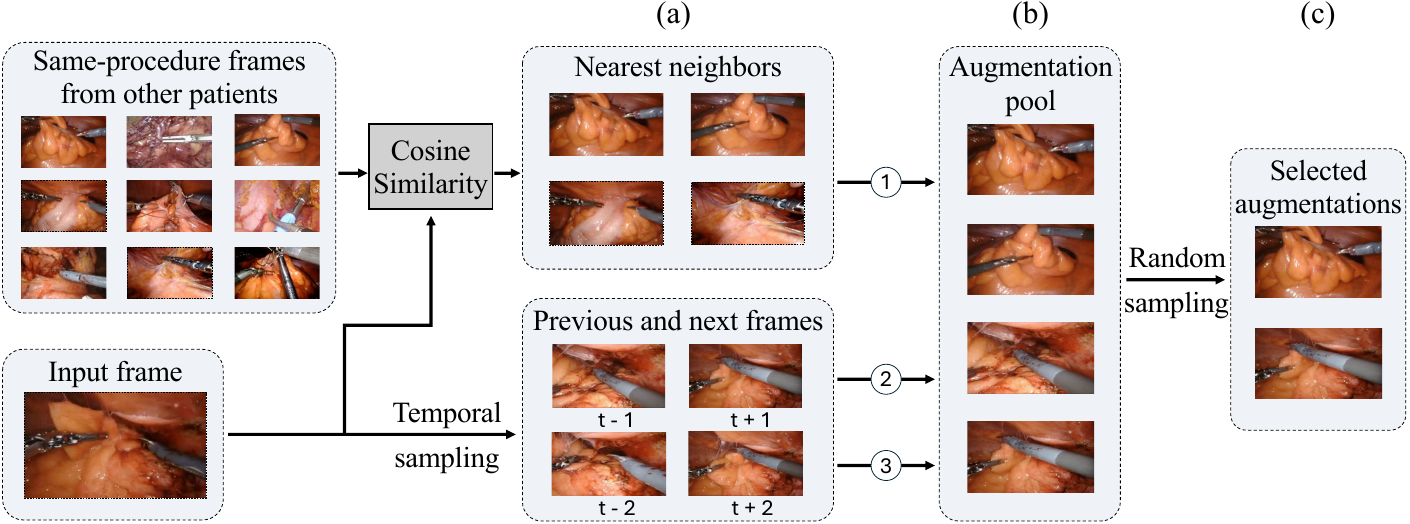}
	\caption{
        \textbf{Proposed augmented distillation method.}
To encourage invariance in LemonFM to minor surgical motion and subtle appearance changes across patients, we introduce $W_i$ (Eq. \ref{eq:loss}).
The key component of $W_i$ is a pair of images (c) that are randomly selected from an augmentation pool (b). 
To populate the augmentation pool, which has capacity for four images, we first retrieve the nearest neighbors$^{(1)}$ of the input image from other videos of the same procedure type, but only include them in the pool if the cosine distance to the input image is smaller than $3\times$ the distance between the input image and its preceding frame in the video (the choice of $3\times$ factor and the cosine distance is justified with ablation experiments in the supplementary material).
When not enough suitable neighbors are found, we supplement the pool with adjacent video frames$^{(2)(3)}$.
    }
	\label{fig:surgical_augmentation}
\end{figure}

\begin{table*}[t]
    \centering
    \small
    \setlength{\tabcolsep}{2.3pt}
    \caption{
        \textbf{Linear probing results for surgical phase recognition, surgical tool presence detection, and surgical action recognition tasks.} The experiments are conducted with freezing the backbone and only fine-tuning a linear classification head to demonstrate the foundation model's effectiveness. The predictions are computed on a frame-by-frame basis for all tasks. Best in \textbf{bold}.
        }
    \label{tab:linear_probing_classification}
    \begin{tabularx}{.9\linewidth}{@{}lcccccccccc@{}}
    \hline
        \multirow{3}{*}{\textbf{Method}} &\multirow{3}{*}{\textbf{Pretraining data}} & \multicolumn{3}{c}{\textbf{Phase recognition }} & ~ & \multicolumn{2}{c}{\textbf{Tool detection }} & ~ & \multicolumn{1}{c}{\textbf{Action recognition}} \\
        \cline{3-5} 
        \cline{7-8} 
        \cline{10-10}
        & ~ &  \textbf{AutoLaparo} & \textbf{Cholec80}& \textbf{M2CAI16}
        & ~ & \textbf{Cholec80} & \textbf{GraSP} 
        & ~ & \textbf{CholecT50} \\
        \cline{3-5} 
        \cline{7-8} 
        \cline{10-10}
        & ~ &  \textbf{Acc/F1} & \textbf{Acc/F1}& \textbf{Acc/F1} 
        & ~ & \textbf{mAP} & \textbf{mAP} 
        & ~ & \textbf{mAP}\\
        \hline
  
        MAE~\cite{He2022MaskedLearners} & LEMON  
        & $35.5/32.0$ & $54.9/43.4$ & $40.7/38.1$ 
        & ~ 
        & $30.6$ & $45.2$ 
        & ~ 
        & $24.8$ 
        \\

        VideoMAEv2~\cite{Wang2023VideoMAEMasking} & LEMON  
        & $49.8/42.4$ & $55.8/48.5$ & $50.7/41.3$ 
        & ~ 
        & $39.6$ & $55.9$ 
        & ~ 
        & $25.8$ 
        \\

        Endo-FM~\cite{Wang2023FoundationPre-train} & Private data 
        & $51.5/43.1$ & $62.7/53.9$ & $53.8/49.8$
        & ~ 
        & $50.0$ & $64.3$
        & ~ 
        & $26.2$
        \\ 
        EndoViT~\cite{Batic2024EndoViT:Images}  & Merged public data 
        &  $45.4/38.7$  &  $46.9/39.0$ & $38.5/37.2$
        & ~  
        & $34.7$  &  $47.1$  
        & ~ 
        &  $29.5$ 
        \\ 
        GSViT~\cite{Schmidgall2024GeneralSurgery} & GenSurgery 
        & $22.0/17.5$ & $20.2/13.5$ & $22.4/18.0$
        & ~ 
        & $18.2$ & $35.6$
        & ~ 
        & $16.5$
        \\ 
        SurgeNetXL~\cite{Jaspers2026ScalingModels}  & SurgeNetXL 
        &  $68.8/57.0$  &  $73.2/65.1$ & $67.8/61.9$
        & ~  
        & $72.1$  &  $62.7$  
        & ~ 
        &  $45.3$  
        \\
        \hline

        LemonFM (\textbf{ours}) & LEMON 
        & $\mathbf{76.4/66.9}$ & $\mathbf{75.8 / 68.6}$ & $\mathbf{68.4/62.4}$
        &~
        & $\mathbf{76.8}$  & $\mathbf{76.4}$ 
        & ~ 
        & $\mathbf{50.4}$ \\
        \hline
    \end{tabularx}
\end{table*}

\noindent
\textbf{Video preprocessing.}
As the videos were from YouTube, we cannot assume that all frames within the trimmed videos are entirely free from non-surgical content (e.g., intraoperative out-of-body views, medical annotations, user-interface details, equipment logos). 
Therefore, we used our trained frame classifier to detect and obliterate the intraoperative non-surgical frames (Fig.~\ref{fig:data_curation_pipeline}d).
Additionally, we manually annotated 2719 surgical frames with 4584 non-surgical bounding-box instances and trained a YOLOv8 model~\cite{Jocher2023UltralyticsYOLO} to detect and obliterate the non-surgical content in surgical frames.
Afterward, we manually quality controlled all curated videos.
We observed that with a $>70\%$ confidence threshold, our automated classifiers achieved $100\%$ precision in identifying surgical videos and eliminated non-surgical frames from those videos with $>99.9\%$ precision.

\noindent
\textbf{Video annotation.}
We cross-referenced video titles with the predefined list of procedures. 
When exact title matches were not found, we leveraged the capabilities of the ChatGPTv4 to perform a more nuanced analysis, incorporating a customized prompt (see supplementary material).
All annotations including the procedure type and the surgery type (robotic or non-robotic) were manually reviewed and validated to ensure the accuracy of assigned labels.

\noindent
\textbf{Ethical Considerations.}
We treat ethical considerations as a key component of our dataset development. 
Our data collection targeted videos from medical institutions and surgeons, aiming to maximize compliance with professional consent standards.
%The data collection process was targeted to collect videos from medical institutions and surgeons, aiming to maximize compliance with professional consent standards.
%
%We collect videos exclusively from YouTube channels run by medical institutions and verified surgeons, which ensures compliance with professional consent standards.
%
While preparing this dataset, two researchers whose PhD specialization is in surgical interventions spent a total of 106 hours annotating and reviewing the data, specifically targeting the removal of out-of-body views, patient identifiers, and other non-surgical content.
%To prepare this dataset, two experts whose PhD specialization is in surgical interventions spent a total of 106 hours annotating and checking the dataset. 
%
%We ensure that we removed all out-of-body views, patient identifiers, and other non-surgical content. 
%
%Because the dataset contains only intraoperative endoscopic views and is anonymized, to the best of our knowledge, patients are not identifiable by current methods. We provide an online form that allows stakeholders to request the removal of their videos from our dataset. Additionally, as videos are linked, if a source video is removed from YouTube, it will be removed from our dataset automatically.
%
Because the dataset consists exclusively of anonymized, intraoperative endoscopic views, patients are, to the best of our knowledge, unidentifiable by current methods. We provide an online form allowing stakeholders to request the removal of their videos. Furthermore, since our dataset links directly to the source material, any video removed from YouTube is automatically removed from our collection.

\subsection{Ground truth and proposed new downstream tasks} 
\label{sec:proposed_tasks}
Each of the 4194 surgical videos is annotated for two tasks: 1) multi-label procedure type classification, where the objective is to identify the procedure[s] performed in an intervention's video from among 35 possible options; and 2) binary surgery type classification, where the goal is to indicate whether a surgery is conducted robotically or non-robotically.
The rationale for these annotations is two-fold. Firstly, it allows us to validate the comprehensiveness of our dataset by ensuring that it encompasses a broad spectrum of surgical procedures. 
Secondly, it facilitates the development of new methods that use this information as a key input (e.g., our LemonFM foundation model, video classification/retrieval, video similarity measurement, and procedure-specific estimation of surgical remaining time).

\section{Proposed LemonFM Foundation Model}
\label{sec:foundation_model}
In this section, we describe how we build LemonFM, our foundation model pretrained on LEMON using a novel augmented distillation approach.
Following the standard practice in the field of surgical computer vision~\cite{Twinanda2017EndoNet:Videos, Wang2022AutoLaparo:Hysterectomy, Nwoye2022Rendezvous:Videos}, we extract frames from the 4194 videos in LEMON at 1 fps for our foundation model pretraining.

\iffalse
\noindent
\textbf{Image preprocessing.}
Surgical videos often exhibit superimposed manufacturer-specific user interface (UI) information generated by the software of the surgical stack.
While certain researchers might derive value from analyzing these UI elements (e.g., some interfaces provide contextual information regarding the instruments utilized in the video), a foundational model for surgery should prioritize learning correlations between tissue appearance and tool-tissue interactions. 
This focus is essential to avoid acquiring spurious correlations with manufacturer-specific UI elements. Therefore, we manually annotated 2719 surgical frames with 4584 non-surgical bounding-box instances and trained a YOLOv8 model \cite{Jocher2023UltralyticsYOLO} to detect and crop out UI content.
\fi

\noindent
\textbf{Proposed augmented distillation method.}
\label{sec:proposed_augmentation}
A foundation model for surgery should learn invariance to 1) minor movements of instruments and tissues observed between adjacent video frames, and 2) minimal appearance changes across patients undergoing the same surgical procedure (e.g., organ color variations). 
This is to encourage the model to produce similar embeddings for frames containing the same surgical tools and exhibiting comparable tissue appearances.

\begin{table*}[t!]
    \centering
    \small
    \setlength{\tabcolsep}{2.9pt}
    \caption{
        \textbf{Full fine-tuning results for surgical phase recognition, surgical tool presence detection, and surgical action recognition tasks.} All experiments are conducted in a full fine-tuning setting (i.e., both backbone and head are fine-tuned). 
Type S denotes a specialist model trained for a specific task, whereas F denotes a foundation model whose generalization is assessed across multiple surgical downstream tasks. 
Marker $^\dag$ indicates the standard 10-second relaxed boundary in phase transitions~\cite{Twinanda2017EndoNet:Videos}, and $^\ddagger$ denotes results from previous studies.
We also report the performance of LemonFM when trained on only 50\% of the labeled training data for the downstream tasks.
}
    \label{tab:finetuned_classification}
    \begin{tabularx}{.9\linewidth}{@{}lcccccccccc@{}}
    \hline
        \multirow{3}{*}{\textbf{Method}} &\multirow{3}{*}{\textbf{Type}} & \multicolumn{3}{c}{\textbf{Phase recognition}} & ~ & \multicolumn{2}{c}{\textbf{Tool detection}} & ~ & \multicolumn{1}{c}{\textbf{Action recognition}} \\
        \cline{3-5} 
        \cline{7-8} 
        \cline{10-10}
        & ~ &  \textbf{AutoLaparo} & \textbf{Cholec80$^\dag$}& \textbf{M2CAI16$^\dag$}
        & ~ & \textbf{Cholec80} & \textbf{GraSP} 
        & ~ & \textbf{CholecT50} \\
        \cline{3-5} 
        \cline{7-8} 
        \cline{10-10}
        & ~ &  \textbf{Acc/Jacc} & \textbf{Acc/Jacc}& \textbf{Acc/Jacc} 
        & ~ & \textbf{mAP} & \textbf{mAP} 
        & ~ & \textbf{mAP}\\
        \hline

        TeCNO$^\ddagger$~\cite{Czempiel2020TeCNO:Networks} & \multirow{3}{*}{S}      
        & $77.3/50.7$ & $88.6/75.1$ & $86.1/74.4$ 
        & ~ 
        & $-$ & $-$ 
        & ~ 
        & $-$ 
        \\

        Trans-SVNet$^\ddagger$~\cite{Jin2022Trans-SVNet:Analysis} & ~   
        & $78.3/50.7$ & $90.3/79.3$ & $87.2/74.7$ 
        & ~ 
        & $-$ & $-$ 
        & ~ 
        & $-$ 
        \\

        LoViT$^\ddagger$~\cite{Liu2025LoViT:Recognition} & ~   
        & $81.4/55.9$ & $92.4/81.2$ & $-/-$ 
        & ~ 
        & $-$ & $-$ 
        & ~ 
        & $-$ 
        \\
        
        \hline

        MAE~\cite{He2022MaskedLearners} &  \multirow{8}{*}{F} 
        & $63.5/42.2$ & $84.6/71.2$ & $77.8/63.6$ 
        & ~ 
        & $72.3$ & $53.0$ 
        & ~ 
        & $33.1$ 
        \\

        VideoMAEv2~\cite{Wang2023VideoMAEMasking} & 
        & $77.4/53.3$ & $87.6/77.2$ & $79.8/70.6$ 
        & ~ 
        & $80.5$ & $78.0$ 
        & ~ 
        & $50.8$ 
        \\

        Endo-FM~\cite{Wang2023FoundationPre-train} &~
        & $80.3/55.3$ & $86.9/76.7$ & $79.1/69.9$
        & ~ 
        & $88.4$ & $84.2$
        & ~ 
        & $54.7$
        \\ 
        EndoViT~\cite{Batic2024EndoViT:Images}  & ~ 
        &  $76.1/54.2$  &  $86.7/75.3$ & $78.5/70.0$
        & ~  
        & $83.9$  &  $77.4$  
        & ~ 
        &  $52.2$ 
        \\ 
        GSViT~\cite{Schmidgall2024GeneralSurgery} & 
        ~
        & $60.5/37.0$ & $75.3/57.8$ & $55.8/39.6$
        & ~ 
        & $34.3$ & $37.7$
        & ~ 
        & $26.3$
        \\ 
        SurgeNetXL~\cite{Jaspers2026ScalingModels}  & ~ 
        &  $85.0^\ddagger/55.1$  & $84.4/72.0$ & $68.5/58.8$
& ~  
        & $86.5$  &  $83.8$  
        & ~ 
        &  $57.5$  
        \\

        LemonFM \textbf{(ours)} - \textit{50\% shot} & ~
        & $84.9/62.4$ & $92.1/78.7$ & $86.1/77.7$
        &~
        & $91.5$  & $89.4$ 
        & ~ 
        & $60.1$ \\

        LemonFM \textbf{(ours)} & ~
        & $\mathbf{85.5/64.8}$ & $\mathbf{92.7/85.1}$ & $\mathbf{89.9/79.4}$
        &~
        & $\mathbf{93.7}$  & $\mathbf{94.4}$ 
        & ~ 
        & $\mathbf{61.9}$ \\
        \hline
    \end{tabularx}
\end{table*}

Inspired by DINO's self-supervised knowledge distillation framework~\cite{Caron2021EmergingTransformers}, we introduce a novel augmented distillation approach that leverages LEMON and employs ConvNeXt-L~\cite{Liu2022A2020s} for both the teacher and student networks.
Specifically, we train a student network $f_{\boldsymbol{\theta_s}}$ to solve for the mapping 
$\sigma(f_{\boldsymbol{\theta_t}} (\boldsymbol{x})) \approx \sigma (f_{\boldsymbol{\theta_s}}(\boldsymbol{x}) )$ 
where 
$\boldsymbol{x} \in \mathbb{R}^{H \times W \times 3}$ 
is an input image, 
$\sigma$ is the softmax function,
$f_{\boldsymbol{\theta_t}} (\boldsymbol{x}) \in \mathbb{R}^C$ 
is a target vector of dimension $C = 2^{16}$ estimated from a teacher network $f_{\boldsymbol{\theta_t}}$. 
The student's parameters $\boldsymbol{\theta_s}$ are updated by gradient descent, while the teacher's parameters $\boldsymbol{\theta_t}$ are updated via exponential moving average from the student.
We use the outputs of both networks to build two categorical distributions, $P_s(z | \boldsymbol{x}) = \sigma_z (f_{\boldsymbol{\theta_s}}(\boldsymbol{x}))$ and $P_t(z | \boldsymbol{x}) = \sigma_z (f_{\boldsymbol{\theta_t}}(\boldsymbol{x}))$, where $\sigma_z$ denotes the $z$-th element of the softmax output, and $z \in \{1, \dots, C\}$. During training, we minimize the following cross-entropy loss between the teacher and student predictions
\begin{align}
    \label{eq:loss}
    \mathcal{L} &= - \sum_{i = 1}^{I} \sum_{\boldsymbol{u} \in U_i} \sum_{\substack{\boldsymbol{v} \in V_i \cup W_i \\ \boldsymbol{u} \neq \boldsymbol{v}}} \sum_{z=1}^C P_t(z|\boldsymbol{u}) \log P_s(z|\boldsymbol{v})
\end{align}
where
$I$ is the number of images, $U_i$ contains two global augmentations of $\boldsymbol{x_i}$, and $V_i$ contains two global augmentations and four augmented crops of $\boldsymbol{x_i}$.
Additionally, we introduce $W_i$ as an extra supervisory signal comprising two images and four augmented crops of these two images.
Fig.~\ref{fig:surgical_augmentation} details the selection procedure, including the neighbor-selection strategy and key parameters.
We use the $k$-nearest neighbors (in embedding space) of the input image from other videos of the same procedure type to learn invariance to appearance changes across patients, and the adjacent video frames of the input image to learn invariance to the motion occurring between consecutive frames.

\noindent
\textbf{Proposed video classification model.}
\label{sec:LemonFM-vid}
Considering that each surgical procedure is localized to a specific region of the body, we posit that a characteristic scene can be associated with each procedure.
Based on this assumption, and inspired by active learning~\cite{Hacohen2022ActiveBudgets}, we propose LemonFM-Vid. This is a video classification head that employs the frame embeddings produced by LemonFM to solve our proposed video classification downstream tasks. It does so by aggregating each frame's embedding following
\begin{align}
    \boldsymbol{v_e} = \sum_{j = 0}^{J - 1} \omega_j \boldsymbol{\phi}_j
    \label{eq:typicality}
\end{align}
where $\boldsymbol{v_e}$ is a video embedding,
$J$ is the number of frames,
$\omega_j = \frac{t(\boldsymbol{\phi}_j)}{\epsilon + \sum_{m = 0}^{J - 1} t(\boldsymbol{\phi}_m)}$,
$\boldsymbol{\phi}_j$ is the embedding of frame $j$,
$t(\boldsymbol{\phi}) = \left( \frac{1}{K} \sum_{\boldsymbol{\eta} \in K-NN(\boldsymbol{\phi})}d(\boldsymbol{\phi}, \boldsymbol{\eta}) \right)^{-1}$ 
is the typicality~\cite{Hacohen2022ActiveBudgets} of a frame,
$K-NN(\boldsymbol{\phi})$ with $K = 20$ are the nearest neighbors of $\boldsymbol{\phi}$,
$d$ is the cosine distance, and $\epsilon = 10^{-8}$.
Then, each $\boldsymbol{v_e}$ is classified with a single-layer MLP.  
 \section{Experiments}
\label{sec:experiments}
\subsection{Downstream Tasks, Datasets and Evaluation}
\label{sec:datasets_and_tasks}
We evaluate LemonFM on four surgical downstream tasks: surgical phase recognition~\cite{Wang2022AutoLaparo:Hysterectomy, Twinanda2017EndoNet:Videos, Stauder2016TheChallenge}, surgical tool presence detection~\cite{Twinanda2017EndoNet:Videos, Ayobi2024Pixel-WiseUnderstanding}, surgical action recognition~\cite{Nwoye2022Rendezvous:Videos}, and surgical semantic segmentation~\cite{Hong2020CholecSeg8k:Cholec80}.
To do so, we utilize widely adopted benchmark datasets sourced from independent hospitals, adhering to established data splits and metrics.
We report official results where available. Otherwise, we independently evaluate models using a unified training protocol to strictly isolate the impact of pretrained weights.

\noindent
\textbf{Surgical phase recognition.} 
For this task, we use three datasets: AutoLaparo~\cite{Wang2022AutoLaparo:Hysterectomy}, M2CAI16~\cite{Twinanda2017EndoNet:Videos}, and Cholec80~\cite{Twinanda2017EndoNet:Videos}.
These datasets include complete surgical workflows with frame-level phase annotations, where each frame corresponds to a single phase.
We perform experiments with both linear probing, where we freeze the backbone and only fine-tune a linear classification head, and full fine-tuning, where both the backbone and the head are fine-tuned. 
We report the frame-wise accuracy and F1-score for the linear probing setting (Table~\ref{tab:linear_probing_classification}), and the video-level accuracy and Jaccard (more details in supplementary material) for the full fine-tuning setting. 
To produce video-level online predictions, we replace the linear head with an MS-TCN head~\cite{Czempiel2020TeCNO:Networks} (Table~\ref{tab:finetuned_classification}).

\noindent
\textbf{Surgical tool presence detection.}
Cholec80 and GraSP~\cite{Ayobi2024Pixel-WiseUnderstanding} are utilized in the surgical tool presence detection task. 
These datasets contain surgical tool presence labels for each frame, allowing multiple tools to be present in a single frame.
We report mean Average Precision (mAP) across all frames for both the linear probing and full fine-tuning settings (see Table~\ref{tab:linear_probing_classification} and~\ref{tab:finetuned_classification}).  
We use a linear head for both experiments.

\noindent
\textbf{Surgical action recognition.}
We use CholecT50~\cite{Nwoye2022Rendezvous:Videos} as the downstream dataset.
Surgical action labels are annotated at frame level, enabling multiple actions occur simultaneously.
We use a linear head for both the linear probing and full fine-tuning settings and apply the same metrics as in the surgical tool presence detection task.

\noindent
\textbf{Surgical semantic segmentation.}
We employ the CholecSeg8k~\cite{Hong2020CholecSeg8k:Cholec80} dataset, which contains pixel-wise semantic annotations for each frame.
The experiments are conducted in full fine-tuning using a feature pyramid network decoder~\cite{Kirillov2019PanopticNetworks}.
Dice score is reported over all frames (Table~\ref{tab:segmentation}).

\begin{table}[t]
    \centering
    \small
    \setlength{\tabcolsep}{5.7pt}
    \caption{
        \textbf{Evaluation on surgical semantic segmentation.} The experiments are conducted with using a feature pyramid network (FPN) decoder~\cite{Kirillov2019PanopticNetworks} in a full fine-tuning manner.
Marker $^*$ denotes a frozen LemonFM backbone, and $^\ddagger$ indicates results from previous studies.
We also report the performance of LemonFM when trained on only 50\% of the labeled training data.
}   
    \label{tab:segmentation}
    \begin{tabularx}{.96\linewidth}{@{}lcc@{}}
    \hline
          \multirow{2}{*}{\textbf{Method}} &  \multirow{2}{*}{\textbf{Pretraining data}}  & \multicolumn{1}{c}{\textbf{CholecSeg8k}}  \\
        \cline{3-3} 
        & &\textbf{mDice}  \\
    \hline

    MAE~\cite{He2022MaskedLearners}  
    & LEMON& 
    $68.5$
 \\
 
    VideoMAEv2~\cite{Wang2023VideoMAEMasking}  
    & LEMON& 
    $65.7$
 \\

    Endo-FM$^\ddagger$~\cite{Wang2023FoundationPre-train} 
    &Private data &
    $56.0$\\
    
    EndoViT$^\ddagger$~\cite{Batic2024EndoViT:Images}
    &Merged public data&
    $71.0$ \\
    
    GSViT$^\ddagger$~\cite{Schmidgall2024GeneralSurgery}
    &GenSurgery&
    $53.0$ \\
    
    SurgeNetXL$^\ddagger$~\cite{Jaspers2026ScalingModels}  
    &SurgeNetXL &
    $69.0$   \\
    \hline

    LemonFM$^*$ 
    &LEMON &
    $71.9$ 
     \\

    LemonFM  - \textit{50\% shot}   
    &LEMON &
    $78.2$  
    
    \\
    
    LemonFM   
    &LEMON &
    $\mathbf{81.3}$  
    
    \\

    \hline
    \end{tabularx}
\end{table}

\subsection{Main Results}
\label{sec:comparision_with_sota}
The performance of our LemonFM compared to the state of the art in each downstream task and dataset is shown in~Fig.~\ref{fig:result_downstreams}, and Tables \ref{tab:linear_probing_classification}, \ref{tab:finetuned_classification} and \ref{tab:segmentation}.
These results demonstrate that the diversity of procedures in LEMON enables LemonFM to effectively extract features that are useful for different surgical procedures and downstream tasks. We provide further experimental results and computational complexity analysis in the supplementary material.

\noindent
\textbf{Surgical phase recognition.}
As shown in Table~\ref{tab:finetuned_classification}, LemonFM outperformed existing surgical foundation models (Endo-FM, EndoViT, GSViT, SurgeNetXL) on AutoLaparo ($>$9pp Jacc.), Cholec80~($>$8pp Jacc.), and M2CAI16~($>$9pp Jacc.). 
It also surpassed state-of-the-art specialist models for these datasets.
The advantage of LemonFM persists under the linear probing setting (Table~\ref{tab:linear_probing_classification}), showing that the pretrained features are inherently informative and transfer well without additional fine‑tuning.

\noindent
\textbf{Surgical tool presence detection.}
As shown in Table~\ref{tab:linear_probing_classification} and \ref{tab:finetuned_classification}, LemonFM outperformed all surgical foundation models in Cholec80 ($>$5pp mAP) and GraSP ($>$10pp mAP).

\noindent
\textbf{Surgical action recognition.}
LemonFM outperformed all surgical foundation models in CholecT50   (Table~\ref{tab:linear_probing_classification} and \ref{tab:finetuned_classification}).

\noindent
\textbf{Surgical semantic segmentation.}
As shown in Table~\ref{tab:segmentation}, LemonFM surpassed every other foundation model in the pixel-wise semantic segmentation task by a large margin ($>$10pp mDice).

\noindent
\textbf{Low data regime.}
As shown in Tables \ref{tab:finetuned_classification} and \ref{tab:segmentation}, when trained on 50\% of the data, LemonFM outperforms all other surgical foundation models fine-tuned on 100\% of the data, demonstrating high data efficiency.

\noindent
\textbf{Robustness evaluation and statistical testing.}
While our proposed LemonFM surpasses other surgical foundation models in each dataset's original single data split, demonstrating the overall strength of the LEMON dataset, we conducted extensive cross-validation to further assess the robustness and stability of LemonFM. We retained the official train/test proportions for each dataset and performed a 5-fold cross-validation under the linear probing protocol. 
As shown in Table~\ref{tab:cross_val}, the low standard deviations across all metrics indicate stable performance across folds and consistent behavior under different data partitions.

\begin{table}[t]
  \centering
  \small
  \setlength{\tabcolsep}{3.2pt}
  \caption{
  \textbf{Cross-validation results.} We report the mean $\pm$ standard deviation (std) across 5-fold cross-validation on each task under linear probing, demonstrating LemonFM's stability.
  }
  \label{tab:cross_val}
  \begin{tabularx}{.96\linewidth}{@{}llcc@{}}
    \hline
    \textbf{Task} & \textbf{Dataset} & \textbf{Metric} & \textbf{Mean $\pm$ Std} \\
    \hline
    Phase recognition & AutoLaparo~\cite{Wang2022AutoLaparo:Hysterectomy} & Acc & 77.9 $\pm$ 2.6 \\
    Tool detection & GraSP~\cite{Ayobi2024Pixel-WiseUnderstanding} & mAP & 70.8 $\pm$ 2.8 \\
    Action recognition & CholecT50~\cite{Nwoye2022Rendezvous:Videos} & mAP & 51.3 $\pm$ 1.4 \\
    Segmentation & CholecSeg8k~\cite{Hong2020CholecSeg8k:Cholec80} & mDice & 72.7 $\pm$ 3.3 \\
    \hline
  \end{tabularx}
\end{table}

\subsection{Ablation Studies}
\label{sec:ablations}
We conducted incremental ablation studies using frame-wise evaluation on phase recognition (AutoLaparo) and semantic segmentation (CholecSeg8k) tasks (Table~\ref{tab:ablations}).

\noindent
\textbf{The significance of LEMON.} 
We compare the effects of using LEMON for pretraining the surgical foundation model against those obtained with general-purpose ImageNet-1K (IN-1K)~\cite{Deng2009ImageNet:Database} and a smaller-scale surgical dataset, Cholec80 (51h)~\cite{Twinanda2017EndoNet:Videos} (rows 1, 2 \& 4, Table~\ref{tab:ablations}). 
Pretraining on LEMON's massive scale (938h) yielded substantial F1-score improvements on AutoLaparo phase recognition over both ImageNet (+12.9pp) and Cholec80 (+19.0pp).
These results confirm that a large-scale, domain-specific dataset like LEMON offers a superior approach for pretraining surgical foundation models.

\noindent
\textbf{The importance of rigorous data curation.}
We analyze the importance of our data curation pipeline (Fig.~\ref{fig:data_curation_pipeline}) by comparing LemonFM against a model pretrained on an uncurated version of LEMON (i.e., the raw surgical YouTube videos without our data curation). 
Our data curation pipeline yields substantial improvements. 
Specifically, the curated dataset leads to an +4.5pp F1-score in the phase recognition task, and +1.3pp mDice in the semantic segmentation task (rows 3 \& 4, Table~\ref{tab:ablations}).

\noindent
\textbf{The effectiveness of the augmented distillation.}
We compared the effectiveness of employing our proposed augmented distillation method (Fig.~\ref{fig:surgical_augmentation}) against a vanilla DINO~\cite{Caron2021EmergingTransformers}. 
LemonFM consistently outperformed the vanilla DINO baseline, with the largest gain (+3.2pp mDice) in semantic segmentation (rows 4 \& 7 in Table~\ref{tab:ablations}).

\begin{table}[t!]
    \centering
    \small
    \setlength{\tabcolsep}{0.5pt}
    \caption{
        \textbf{Ablation studies.}
        We froze the backbone and only fine-tuned the task-specific heads (i.e., a linear head for phase recognition and a FPN head for semantic segmentation).
The column ``Curation'' marks whether the raw surgical videos were processed with the pipeline in Fig.~\ref{fig:data_curation_pipeline}. The column “Aug” shows whether the augmented distillation method in Fig.~\ref{fig:surgical_augmentation} was applied.
    }
    \label{tab:ablations}
    \begin{tabularx}{\linewidth}{@{}lcccccc@{}}
    \hline
        \multirow{2}{*}{\textbf{Dataset}}  & \multirow{2}{*}{\textbf{Curation}}  & \multirow{2}{*}{\textbf{Aug}}  &  \multirow{2}{*}{\textbf{Backbone}}  & \multicolumn{1}{c}{\textbf{AutoLaparo}}  &  ~  & \multicolumn{1}{c}{\textbf{CholecSeg8k}} 

        \\
        \cline{5-5} 
        \cline{7-7} 
        &  & & & \textbf{Acc/F1}  
        &  ~  &  \textbf{mDice} 
        \\
    \hline 
    
     IN-1K
        & N/A 
        & \ding{55} 
        & ConvNeXt-L
        & $63.6/53.0$  
        & ~
        &  $64.4$   \\

    Cholec80
        & N/A 
        & \ding{55} 
        & ConvNeXt-L
        & $54.0/46.9$  
        & ~
        &  $64.1$   \\
    
    LEMON
        & \ding{55} 
        & \ding{55} 
        & ConvNeXt-L
        &  $71.7/61.4$  
        &  ~
        &  $67.4$  
    \\
    LEMON
        & \checkmark 
        & \ding{55} 
        & ConvNeXt-L
        &  $75.3/65.9$  
        &  ~
        &  $68.7$  
    \\

    LEMON
        & \checkmark 
        & \checkmark  
        & ViT-L
        &  $75.6/66.1$  
        &  ~
        &  $61.2$  
    \\
    LEMON
        & \checkmark 
        & \checkmark  
        & ConvNeXt-B
        &  $74.7/64.5$  
        &  ~
        &  $68.5$  
    \\
    
    LEMON
        & \checkmark 
        & \checkmark 
        & ConvNeXt-L
        &  $\mathbf{76.4/66.9}$  
        &  ~
        &  $\mathbf{71.9}$  
    \\
    
    \hline

    \end{tabularx}
\end{table}

\noindent
\textbf{LemonFM design choices.} 
%We evaluate the effects of two design choices for surgical foundation model pretraining: backbone architecture, ViT vs. ConvNeXt, and self-supervision method, generative-based~\cite{He2022MaskedLearners} vs. discriminative-based~\cite{Caron2021EmergingTransformers}.
We evaluate the effects of three key design choices for surgical foundation model pretraining: backbone architecture (ViT vs. ConvNeXt), self-supervision method (generative-based~\cite{He2022MaskedLearners} vs. discriminative-based~\cite{Caron2021EmergingTransformers}), and model type (image-based vs. video-based).
As shown in rows 5 and 7 of Table~\ref{tab:ablations}, when pretrained on LEMON with our augmented knowledge distillation, the \textbf{ConvNeXt backbone} consistently outperformed the \textbf{ViT-based counterpart}.
%
%when both pretrained on the curated LEMON dataset with the augmented knowledge distillation, the one using ConvNeXt as backbone outperforms the ViT‑based counterpart, 
The improvement in performance is particularly notable on the pixel-wise semantic segmentation task (+10.7pp mDice). %Moreover, the large ConvNeXt variant surpasses the base model by an additional margin.
We hypothesize this is because ConvNeXt's hierarchical structure and convolutional inductive biases (e.g., local connectivity) preserve fine-grained surgical details like tool tips, which ViT's early patchification collapses~\cite{Yuan2025SurgicalEnhancement, Liu2022A2020s}.
In Tables~\ref{tab:linear_probing_classification}, \ref{tab:finetuned_classification}, and~\ref{tab:segmentation}, we compare methods whose pretext task is \textbf{generative-based} (MAE, VideoMAEv2, EndoViT) against methods whose pretext task is \textbf{discriminative-based} (classification) such as Endo-FM, SurgeNetXL and our LemonFM. 
%
%
%MAE and EndoViT are MAE-based, whereas 
%Endo‑FM, SurgeNetXL, and our LemonFM are based on discriminative-based method.
%
%
%MAE and EndoViT employ generative-based method, whereas Endo‑FM, SurgeNetXL, and our LemonFM are based on discriminative-based method.
%
The results show that discriminative‑based models yield a substantial performance advantage, especially when the backbone is frozen, indicating generative objectives such as MAE are less effective at capturing surgical visual cues.
Finally, we justify the choice of an \textbf{image-based} foundation model over a \textbf{video-specific} one like VideoMAEv2~\cite{Wang2023VideoMAEMasking}. Although VideoMAEv2 is pretrained on spatio-temporal data, it transfers poorly to dense image tasks, as seen by its 65.7\% mDice score on segmentation (Table~\ref{tab:segmentation}), a significant drop of 15.6pp compared to our LemonFM. 
More importantly, using a pure image encoder (LemonFM) followed by a thin temporal TCN head~\cite{Czempiel2020TeCNO:Networks} allows for real-time performance ($\le 40$ms/frame~\cite{Stenmark2022Vision-BasedSurgery, Ceron2022Real-timeFusion}) at $10$ms/frame, far surpassing the $65$ms/frame required by VideoMAEv2 (ViT-B/16), which is critical for real-time surgical applications. 
Overall, these results support our design of a superior and highly efficient foundation model capable of handling both image and video downstream tasks.

\begin{table}[t!]
    \centering
    \small
    \setlength{\tabcolsep}{2.8pt}
    \caption{
        \textbf{Leaderboard for the proposed tasks on LEMON.}
Multi-label (35 classes) video classification of procedure types and binary classification of surgery types. 
}
    \label{tab:leaderboard}
    \begin{tabularx}{\linewidth}{@{}lccc@{}}
    \hline
        \multirow{2}{*}{\textbf{Method}} 
& \multicolumn{1}{c}{\textbf{Procedure type}} 
        & ~ 
        & \multicolumn{1}{c}{\textbf{Surgery type}}  \\
        \cline{2-2}
        \cline{4-4}
& \textbf{mAP/F1} 
        & ~ 
        & \textbf{Acc/F1}
        \\
    \hline 

    SlowFast~\cite{Feichtenhofer2019SlowFastRecognition} 
& 22.0 /23.9 
    & ~
    & 88.5 /87.5     
    \\
    TimeSformer~\cite{Bertasius2021IsUnderstanding}  & 42.1/37.5 
    & ~
    & 93.2 /92.7       
    \\

    MViTv2~\cite{Li2022MViTv2:Detection}   & 49.5/41.8
    & ~
    & 95.8 /94.6     
    \\
    
    Video Swin Transformer~\cite{Liu2022VideoTransformer}   & 51.4/47.9 
    & ~
    & 98.8/98.7        
    \\
    
    \hline
    LemonFM-Vid            
& $\mathbf{57.8/49.3}$ 
    & ~
    & $\mathbf{98.9/98.9}$ 
    \\

    \hline
    \end{tabularx}
    \vspace{-0.7em}
\end{table}

\subsection{Leaderboard}
\label{sec:leaderboard}
We benchmarked state-of-the-art video classification models on LEMON (Table~\ref{tab:leaderboard}), including SlowFast~\cite{Feichtenhofer2019SlowFastRecognition}, TimeSformer~\cite{Bertasius2021IsUnderstanding}, MViTv2~\cite{Li2022MViTv2:Detection} and Video Swin Transformer~\cite{Liu2022VideoTransformer}.
Leveraging the robust representations extracted from LemonFM, our video classification model, LemonFM-Vid (Sec.~\ref{sec:LemonFM-vid}), improved multi‑label procedure classification by 6.4pp mAP over all leading models and matched the top performance of the Video Swin Transformer on the binary surgery type classification task.
The incorporation of frame-level features extracted by LemonFM, in conjunction with the weighted feature embeddings informed by our frame typicality (Eq.~\ref{eq:typicality}), led to substantial performance gains, underscoring the value of LemonFM for surgical video analysis tasks.
Nonetheless, the mAP score of 57.8\% achieved by LemonFM-Vid (the top performing method) highlights the challenges associated with our newly proposed procedure type video classification task.
Our review of LemonFM-Vid predictions shows these confusions are not random, but primarily occur between anatomically adjacent procedures or those that share instruments, such as myomectomy and hysterectomy. This underscores the need for continued innovation to resolve these specific visual ambiguities.

\vspace{-0.2em}

 \section{Conclusion}
\label{sec:conclusion}
Building on a large corpus of publicly available surgical videos, in this work we presented three main contributions: 1) we proposed a novel data curation pipeline to curate online surgical videos; 2) leveraging this pipeline, we created a large-scale and diverse surgical dataset, LEMON; and 3) we developed a foundation model specifically for surgical applications, LemonFM.
Our results show that LemonFM outperformed both state-of-the-art surgical foundation models and task-specific architectures across diverse surgical downstream tasks. 
Even when fine-tuned with only half of the labeled data, LemonFM maintained its advantages. 
These findings demonstrate LemonFM's effectiveness and verify the quality of LEMON as a valuable resource for computer vision research in surgery.
We expect that both LEMON and LemonFM will serve as a foundation for future research in surgical computer vision and facilitate the advancement of autonomous robotic surgery systems.
Future work will concentrate on the development of a surgery-specific video foundation model.

\appendix
\clearpage
\setcounter{page}{1}
\maketitlesupplementary

\begin{figure*}[t!]
	\centering
	\includegraphics[width=\linewidth]{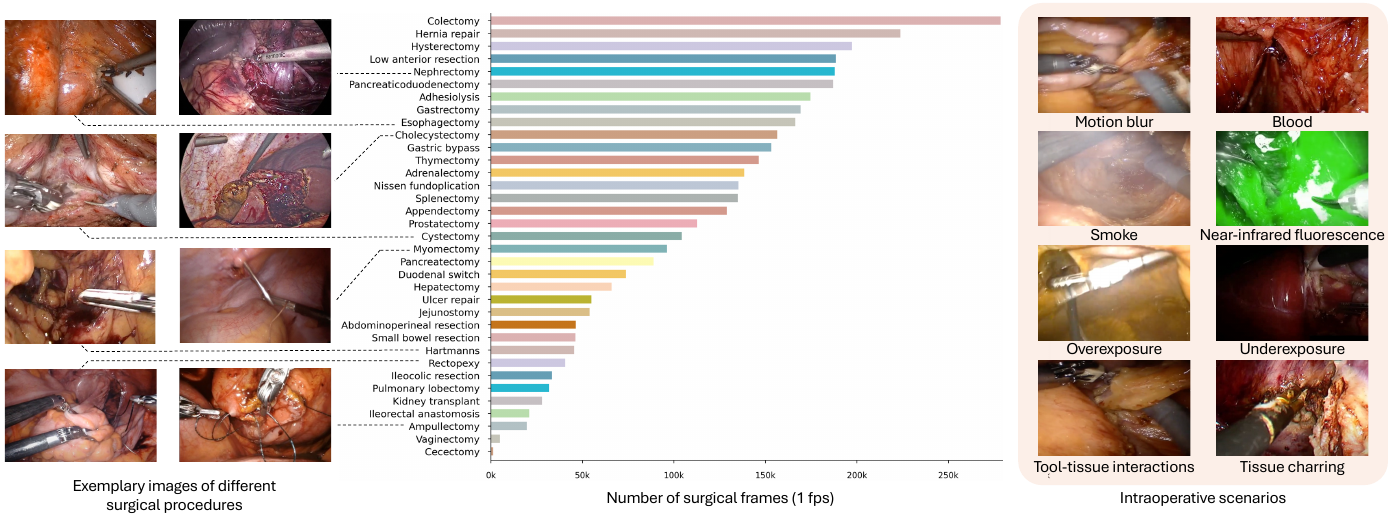}
	\caption{
        \textbf{Diversity and procedure prevalence in LEMON.}
Representative samples from various procedures, demonstrating the diverse range of cases in our curated dataset (left, right).
Distribution of surgical frames by procedure type (center).
    }
	\label{fig:statistic_dataset}
\end{figure*}

\section{Surgical Background}
\label{sec:surgical_background}
In contrast to traditional open surgery (not the target of our work), which entails extensive tissue disruption via large incisions and relies on direct visual inspection by the surgeon (i.e., no cameras are used), modern surgical practices predominantly employ minimally invasive techniques (the target of our work). 
These techniques involve the insertion of slender instruments, including one that is a camera, into the patient's body through small incisions.

To control the instruments from outside the patient's body, there are two variants: robotic-assisted surgery (called ``robotic'' in our manuscript) and non-robotic laparoscopy (called ``non-robotic'' in our manuscript).
Robotic-assisted surgery involves a surgeon sitting on a console and controlling (with joysticks) several robotic arms that hold and steer the instruments inside the patient. 
In contrast, non-robotic (also known as traditional or conventional) laparoscopy requires the surgeon to directly hold and steer the laparoscopic hand-held instruments (i.e., no robotic arms involved).

\section{Data Curation Details}
\label{sec:details_datacuration}
This section details our data curation pipeline and presents the procedure diversity and distribution of LEMON~(Fig.~\ref{fig:statistic_dataset}).
From an initial pool of 18K raw videos, our video classification filtering retained 6617. Subsequent trimming and preprocessing removed 66 hours of non-surgical footage, yielding the final LEMON dataset of 4194 videos (938 hours).
% This section provides details of our data curation pipeline and presents LEMON’s procedure diversity and prevalence~(Fig.~\ref{fig:statistic_dataset}).
% %
% From 18K raw videos, video classification retained 6617. Trimming and preprocessing removed 66h of non-surgical footage, yielding 4194 LEMON videos (938h).

\subsection{Video Classification}

\textbf{Video summarization.}
To improve the efficiency of annotating surgical and non-surgical videos, we obtained 4$\times$4-image video storyboards (i.e., a single image containing a collage of key video frames) for all the collected videos using the method described in \cite{Garcia-Peraza-Herrera2023VideoSum:Summarization}. 
Storyboards enabled us to quickly determine whether a video contained substantial surgical footage, thereby avoiding the need for complex analysis on the entire sequence.

\noindent
\textbf{Video storyboard classification.}
We manually annotated a dataset comprising 2160 surgical and 1910 non-surgical storyboards.
The annotation criterion adopted for labeling a storyboard as \textit{surgical} was that at least 50\% of the key frames contained shots from a surgical camera; specifically, open surgery videos were categorized as non-surgical.
To classify the rest of the collected videos, we trained a ResNet18~\cite{He2016DeepRecognition}.
To ensure the accuracy of the inference results, we manually reviewed all the videos classified as surgical.

\noindent
\textbf{Performance of video storyboard classification models.}
We trained five video storyboard classification models (different data splits) to categorize the videos as either surgical or non-surgical.
Each fold was split into training, validation, and testing sets with ratios of 0.8, 0.1, and 0.1, respectively.
The average F1-score of the video storyboard classification models was $94.75\%\pm1.1$. 
The results for each cross-validation fold are shown in Table~\ref{tab:video_summary_performance}.

\begin{table}[t]
\centering
\small
\setlength{\tabcolsep}{1.2pt}
\caption{Performance of the video storyboard classification models across five folds.}
\label{tab:video_summary_performance}
\begin{tabularx}{\linewidth}{@{}lcccc@{}}  \toprule
    \textbf{Fold} & \textbf{F$_1$-score (\%)} & \textbf{Accuracy (\%)} & \textbf{Precision (\%)} & \textbf{Recall (\%)} \\
    \midrule
    Fold 0 & 94.42 & 94.50 & 95.88 & 93.00 \\  
    Fold 1 & 93.60 & 93.50 & 92.23 & 95.00 \\ 
    Fold 2 & 93.89 & 93.75 & 91.87 & 96.00 \\ 
    Fold 3 & 96.50 & 96.50 & 96.50 & 96.50 \\ 
    Fold 4 & 95.36 & 95.50 & 98.40 & 92.50 \\ 
    \midrule
    Average & 94.75 & 94.75 & 94.98 & 94.60 \\ 
    Std Dev & 1.10 & 1.16 & 2.66 & 1.57 \\
    \bottomrule
\end{tabularx}
\end{table}

\subsection{Video Selection and Trimming}
\textbf{Frame classification.}
A ResNet18~\cite{He2016DeepRecognition} was trained for the surgical/non-surgical video frame classification task.
To produce the annotations, the videos were sampled at one frame per second (fps).
We annotated 7967 frames, 5481 of which turned out to be surgical and 2486 non-surgical.

\noindent
\textbf{Video trimming.}
Most online videos contain introductory and conclusion slides. 
Experimentally, we found that the start and end of the surgical footage can be reliably identified by finding the first and last three consecutive frames classified as surgical by our surgical frame classifier (sampling the video at 1\,fps). 
Therefore, this is the approach we used to discard the non-surgical parts at the beginning and end of the collected videos. 
The resulting videos were manually quality checked.

\noindent
\textbf{Performance of frame classification models.}
We trained five frame classification models to classify a video frame as either surgical or non-surgical.
Each model of the five was trained in a different training-validation-testing split of the data, with a split rate of 0.8, 0.1, and 0.1.
As shown in Table~\ref{tab:classification_models_surgical_and_non-surgical}, the average F1-score of the frame classification models was $95.64\% \pm0.94$.

\subsection{Video Preprocessing}
We use our trained frame classifier to detect and remove the intraoperative non-surgical frames.
Additionally, we manually annotated 2719 surgical frames with 4584 non-surgical bounding-box instances and trained a YOLOv8 Nano model~\cite{Jocher2023UltralyticsYOLO} to detect and obliterate the non-surgical content in surgical frames.
The positions of the non-surgical content bounding boxes that have been obliterated from the video frames in LEMON are provided in a JSON file for those researchers who wish to know where the coordinates of the non-surgical information (e.g., UI elements containing instrument names) are located in the original frames.
After this process, we manually quality controlled all the remaining curated videos.

\noindent
\textbf{Performance of non-surgical content detection models.}
We trained five non-surgical content detection models to detect and obliterate \textit{non-surgical} regions in \textit{surgical} video frames.
Each model was trained on a different training-validation-testing split of the data, with a split ratio of 0.8, 0.1, and 0.1, respectively.
The average mAP50 of the five models was $79.29\% \pm 3.4$, and the average mAP50-95 was $66.18\% \pm 2.98$. 
The results for all the cross-validation folds are shown in Table~\ref{tab:object_detection_tesing}.

\begin{figure*}[ht]
\centering
\begin{lstlisting}
You are a highly knowledgeable assistant specializing in surgical procedures and medical terminology. 
Your expertise includes identifying and categorizing surgical interventions based on clinical descriptions and procedural contexts.
Here is a list of 35 possible surgical procedure types: pancreatectomy, pancreaticoduodenectomy, splenectomy, ampullectomy, hepatectomy, nephrectomy, low anterior resection, colectomy, abdominoperineal resection, pulmonary lobectomy, hartmanns, prostatectomy, gastric bypass, duodenal switch, gastrectomy, small bowel resection, hernia repair, ulcer repair, cholecystectomy, appendectomy, ileocolic resection, cecectomy, myomectomy, hysterectomy, nissen fundoplication, adrenalectomy, thymectomy, rectopexy, adhesiolysis, esophagectomy, cystectomy, jejunostomy, ileorectal anastomosis, kidney transplant, vaginectomy. 
Based on the description of the surgical video: <video title>, determine the most likely procedure type from the list. Focus on matching the description to the procedure type that best aligns with the terminology and context provided.
\end{lstlisting}
\caption{
    ChatGPT prompt employed to match video titles to procedure types.
The title of the video to be matched is inserted where the $<$video title$>$ tag is located. 
}
\label{fig:gpt_prompt}
\end{figure*}

\subsection{Video Annotation}
For the surgery type, a video is considered to be robotic if the video title includes any of the following keywords: \textit{Robotic}, \textit{Robot}, 
\textit{Robo},
\textit{Hugo}, \textit{Versius}, \textit{Senhance}, \textit{Telerobotic}, \textit{Console}, and \textit{da Vinci}. 
The search for these terms in the video titles was case-insensitive.
The remaining videos, with titles that do not include any of these keywords, were manually verified to ensure that they are manual surgical procedures.
For the surgical procedure type, we cross-reference the video titles with the predefined list of procedures. 
When exact title matches are not found, we leverage the capabilities of the ChatGPTv4 API to perform a more nuanced analysis, incorporating a customized prompt as shown in Fig.~\ref{fig:gpt_prompt}.
All annotations were manually quality controlled.
%After curation, the final LEMON dataset resulted in $4194$ surgical videos comprising 35 distinct procedure types, 94\% of which are $1280\times720p$, the remaining $6\%$ have varied resolutions, with a minimum of $640\times480$ pixels. 

After curation, the final LEMON dataset consisted of $4194$ surgical videos across $35$ distinct procedure types, with 94\% in $1280\times720$ resolution and the remaining 6\% in varied resolutions, the smallest being $640\times480$ pixels.

\begin{table}[t]
\centering
\small
\setlength{\tabcolsep}{1.2pt}
\caption{Performance of frame classification models across five folds.}
\label{tab:classification_models_surgical_and_non-surgical}
\begin{tabularx}{\linewidth}{@{}lcccc@{}} \toprule
    \textbf{Fold} & \textbf{F$_1$-score (\%)} & \textbf{Accuracy (\%)} & \textbf{Precision (\%)} & \textbf{Recall (\%)} \\
    \midrule
    Fold 0 & 94.65 & 94.85 & 98.37 & 91.21 \\
    Fold 1 & 96.43 & 96.48 & 97.93 & 94.98 \\
    Fold 2 & 96.53 & 96.61 & 98.69 & 94.47 \\
    Fold 3 & 94.44 & 94.60 & 97.33 & 91.71 \\
    Fold 4 & 96.15 & 96.23 & 98.17 & 94.22 \\
    \midrule
    Average & 95.64 & 95.75 & 98.10 & 93.32 \\
    Std Dev & 0.94 & 0.88 & 0.49 & 1.39 \\
    \bottomrule
\end{tabularx}
\end{table}

\subsection{Curation Precision Strategy}
To ensure high precision during dataset curation, we enforced strict filtering criteria. We utilized an ensemble of five models derived from 5-fold cross-validation for our three curation components: video storyboard classification, video frame classification, and non-surgical region detection. During inference, each model applied a confidence threshold of 70\%. Final predictions were determined via majority voting (for storyboard and frame classification) and non-maximum suppression (for region detection). This rigorous approach proved highly effective: our final review confirmed that the pipeline achieved 100\% precision at the video level and $>99.9\%$ precision at the frame level.

For the annotation verification, we confirmed that the automated pipeline generated procedure-type labels (e.g., thymectomy, cystectomy) with 95.2\% accuracy, and surgery-type labels (e.g., robotic, non-robotic) with 97.6\% accuracy. The experts subsequently corrected all identified errors.

\subsection{Human Effort}
We provide a detailed breakdown of manual effort (in person-hours). The curation workload comprises labeling 4K storyboards (3 h) and 8K frames (3 h), followed by annotating 2719 surgical frames with 4584 non-surgical bounding boxes (10 h). The most extensive process involved manually reviewing LEMON videos, which were previously curated by models trained on the initial annotations, to confirm surgical content and remove artifacts such as out-of-body views (72 h), and finally verifying video annotations for surgery and procedure types (18 h).

\subsection{Data Quality and Reliability}
To validate LEMON's labels, two researchers (Chengan Che and Chao Wang) independently annotated a stratified random subset of \textbf{500 videos ($>$10\%)}, balanced across 35 procedure types.
We achieved a \textbf{Cohen's Kappa} of \textbf{0.97} for procedure types and \textbf{$>$0.99} for surgery types, confirming robust label reliability.
This high agreement reflects the clear visual distinctiveness of the procedures and the rich original metadata provided by verified medical professionals.

\section{LemonFM Pretraining}

\subsection{Pretraining Details}
For LemonFM pretraining, we trained on an Ubuntu 22.04.5 LTS node with eight NVIDIA V100 GPUs (32GB each), using a batch size of 24 per GPU under PyTorch 2.5.1+cu124 (CUDA 12.4).
We employed AdamW as the optimizer with a teacher temperature of 0.04, fp16 precision, an initial learning rate of 5e-4 (after warm-up), a minimum learning rate of 1e-6, and a random seed of 30.
The model was trained for 60 epochs, including 10 warm-up epochs. 
This number of epochs enables the model to converge and stabilize its training loss on our dataset and pretext task.
The model with the lowest training loss was selected as the final model.

\subsection{Augmented Distillation Design Choices}

\noindent
\textbf{Cosine Similarity vs. L2 distance.}
%
%We use cosine similarity for neighbor mining, following standard SSL practice (e.g., SwAV~\cite{Caron2020UnsupervisedAssignments}, NNCLR~\cite{Dwibedi2021WithRepresentations}). We validated this choice against L2 (Euclidean) distance using a k-NN evaluation for phase recognition. In this test, labels were predicted by a majority vote of the 20 nearest training-set neighbors. Cosine similarity significantly outperformed L2 distance, yielding accuracy gains of 4 pp (71.3 to 75.3) on AutoLaparo and 3.5 pp (66.8 to 70.3) on Cholec80.
%
We use cosine similarity for neighbor mining, following standard SSL practice (e.g., SwAV~\cite{Caron2020UnsupervisedAssignments}, NNCLR~\cite{Dwibedi2021WithRepresentations}). We validate this choice against L2 (Euclidean) distance using a \textit{k}-NN evaluation for phase recognition, where labels are predicted by a majority vote over the 20 nearest training-set neighbors. 
To isolate the impact of the distance metric, these experiments were conducted using a ConvNeXt-L~\cite{Liu2022A2020s} backbone pretrained with the vanilla DINO method~\cite{Caron2021EmergingTransformers}, excluding our proposed surgical augmentations. 
Results show that cosine similarity significantly outperforms L2 distance, yielding accuracy gains of 4 pp (71.3 to 75.3) on AutoLaparo and 3.5 pp (66.8 to 70.3) on Cholec80.

\noindent
\textbf{Sensitivity to augmented distillation threshold.}
We further analyze the impact of the cosine distance threshold for our augmented distillation (Fig.~4 in the main manuscript). To perform this assessment in a computationally efficient manner, we employ a ConvNeXt-S backbone.
Linear probing on AutoLaparo phase recognition yields Acc/F1 scores of 72.9/63.2 ($1.5\times$), \textbf{73.4/63.7 ($3\times$)}, and 72.6/63.0 ($6\times$). Similarly, Cholec80 phase recognition yields Acc/F1 scores of 72.2/65.2 ($1.5\times$), \textbf{73.6/66.7 ($3\times$)}, and 72.7/65.2 ($6\times$).
These results confirm that our $3\times$ threshold is near-optimal. Qualitatively, a stricter 1.5$\times$ threshold selects too few cross-video neighbors (reducing diversity), while a looser 6$\times$ threshold admits dissimilar, noisy views that slightly degrade performance.

\iffalse
\begin{figure}[t!]
	\centering
	\includegraphics[width=\linewidth]{obliteration0305.pdf}
	\caption{
        \textbf{Image preprocessing in LemonFM.}
We automatically detect and crop out the user interface information from the surgical frames.
    }
	\label{fig:obliteration}
\end{figure}
\fi

\section{Downstream Task Details}
\label{sec:dataset_split}
In this section, we provide details on the datasets used for evaluating each downstream task, including their respective data splits, as well as training configurations for linear probing and full fine-tuning settings.
The experiments were conducted on an Ubuntu 22.04.5 LTS node equipped with an Intel IceLake Xeon CPU (72 vCPUs) and two NVIDIA A100 GPUs (each with 80 GB of memory), using PyTorch 2.5.1+cu124 (CUDA 12.4).

\subsection{Data Splits}
\label{sec:downstream_data_splits}
\textbf{Surgical phase recognition.} 
For AutoLaparo, we followed the standard split of ten training videos, four for validation, and seven for testing~\cite{Wang2022AutoLaparo:Hysterectomy}.
For M2CAI16, we followed the data splits in~\cite{Jin2018SV-RCNet:Network, Twinanda2017EndoNet:Videos,Jin2022Trans-SVNet:Analysis}, dividing the dataset into 27 training videos and 14 testing videos, respectively.
For Cholec80, we adopted the data splits in~\cite{Twinanda2017EndoNet:Videos, Czempiel2020TeCNO:Networks, Liu2025LoViT:Recognition, Jin2020Multi-taskAnalysis}, allocating 40 videos for training, eight for validation, and 32 for testing.

\begin{table}[t!]
\centering
\small
\setlength{\tabcolsep}{1.5pt}
\caption{Performance of bounding box detection models across five folds.}
\vspace{5pt}
\label{tab:object_detection_tesing}
\begin{tabularx}{\linewidth}{@{}lcccccc@{}}
    \toprule
    \textbf{Fold} & \textbf{Images} & \textbf{Instances} & \textbf{Precision} & \textbf{Recall} & \textbf{mAP50} & \textbf{mAP50-95} \\
    \midrule
    Fold 0 & 272 & 447 & 74.79 & 77.63 & 75.50 & 62.00 \\
    Fold 1 & 272 & 438 & 70.73 & 77.40 & 76.36 & 61.79 \\
    Fold 2 & 272 & 509 & 83.34 & 79.60 & 84.89 & 68.56 \\
    Fold 3 & 272 & 422 & 75.27 & 83.69 & 80.31 & 68.98 \\
    Fold 4 & 272 & 496 & 81.32 & 81.45 & 82.58 & 68.65 \\
    \midrule
    Average & -- & -- & 77.09 & 79.94 & 79.29 & 66.18 \\
    Std Dev& -- & -- & 4.41 & 2.33 & 3.40 & 2.98 \\
    \bottomrule
\end{tabularx}
\end{table}

\noindent
\textbf{Surgical tool presence detection.} 
For GraSP, we adopted the data splits specified in~\cite{Ayobi2024Pixel-WiseUnderstanding}, with four videos for training, four for validation, and five for testing.
The data split for Cholec80 is identical to those employed in the surgical phase recognition task.

\noindent
\textbf{Surgical action recognition.} For CholecT50, we followed the formal data split as proposed in~\cite{Nwoye2022Rendezvous:Videos}.

\noindent
\textbf{Surgical semantic segmentation.} For CholecSeg8k, we followed previous works~\cite{Jaspers2026ScalingModels,Grammatikopoulou2023AVideos} and used 75\% of the videos for training and 25\% of the videos for testing (videos 12, 20, 48 and 55).

\subsection{Linear Probing}
\label{sec:downstream_details_linear_probing}
We used the teacher model from LemonFM as our backbone. 
The images were resized to $224\times224$ for all downstream tasks.
The downstream models with linear heads were trained with a batch size of 512, an initial learning rate (LR) of 1e-3, the AdamW optimizer, a random seed of 30, and cross-entropy loss. 
An early stopping criterion was applied, stopping training after 10 epochs if the validation loss shows no improvement.

Furthermore, we report video-level linear probing results to compare LemonFM against SurgeNetXL, a leading surgical foundation model (Table~\ref{tab:video_linear_probing}). 

\begin{table}[t!]
    \centering
    \small
    \setlength{\tabcolsep}{5.1pt}
    \caption{
        Video-level linear probing results. Performance of LemonFM is compared against SurgeNetXL across three surgical datasets. Metrics are reported as Accuracy/F1-score. % Update these metric names if they differ from Acc/F1
    }
    \label{tab:video_linear_probing}
    \begin{tabularx}{.95\linewidth}{@{}lccc@{}}
    \toprule
    \textbf{Method} & \textbf{AutoLaparo} & \textbf{Cholec80} & \textbf{M2CAI16} \\
    \midrule
    SurgeNetXL~\cite{Jaspers2026ScalingModels} & 74.0/50.7 & 73.4/51.9 & 64.5/42.9 \\
    LemonFM & $\mathbf{76.9/53.8}$ & $\mathbf{75.1/52.7}$ & $\mathbf{66.7/44.1}$ \\
    \bottomrule
    \end{tabularx}
\end{table}

\subsection{Full Fine-tuning}
\label{sec:downstream_details_finetuning}
We used the teacher model from LemonFM as our backbone.
The images were resized to $224\times224$ for all downstream tasks.
The downstream models were trained with a batch size of 112, an initial learning rate (LR) of 1e-4, the AdamW optimizer, a random seed of 30, and cross-entropy loss. 
We used five-fold cross-validation for 50\% shot fine-tuning.
An early stopping criterion was applied, stopping training after 10 epochs if the validation loss shows no improvement.

\noindent
\textbf{Surgical phase recognition.}
To compare with other specialist models that are specifically tailored for this task, we adopt the video-level accuracy and Jaccard in the full fine-tuning setting. 
The accuracy for a video is computed by dividing the number of frames whose class has been correctly predicted by the total number of frames.
The overall accuracy for a dataset is defined as the mean accuracy value over all videos, henceforth referred to as \textit{video-level} accuracy \cite{Jin2021EvaluationVideo, Jin2021TemporalVideo, Liu2025LoViT:Recognition, Jin2020Multi-taskAnalysis}.
The Jaccard is computed for each class and video, then averaged: first within videos, and secondly across classes~\cite{Jin2021EvaluationVideo, Jin2021TemporalVideo, Liu2025LoViT:Recognition, Jin2020Multi-taskAnalysis}.

During training, we utilized a TCN~\cite{Czempiel2020TeCNO:Networks} head and followed the two-stage training approach outlined in TeCNO~\cite{Czempiel2020TeCNO:Networks}.
In the first stage, we trained the backbone with a linear head to perform frame-wise classification without temporal context. 
Then, we employed the TCN head to incorporate temporal information of the extracted features for predictions. The default TCN configuration was used, consisting of two stages, each containing nine layers. 
The TCN head was trained using the Adam optimizer with an initial learning rate of 5e-3 for 70 epochs to ensure sufficient training. 
The checkpoint with the lowest validation loss was selected as the optimal checkpoint.

\begin{table}[t!]
    \centering
    \small
    \setlength{\tabcolsep}{5.1pt}
    \caption{
        Computational cost analysis.
        Inference and training times are measured in milliseconds (ms). Memory usage is reported in Megabytes (MB).
    }
    \label{tab:compute_cost}
    \begin{tabularx}{1.0\linewidth}{@{}lcccccc@{}}
    \hline
        \multirow{2}{*}{\textbf{Model}} & \multicolumn{2}{c}{\textbf{Inference}} & &\multicolumn{2}{c}{\textbf{Training}} & \multirow{2}{*}{\textbf{Acc (\%)}} \\
        \cline{2-3} \cline{5-6} 
         & Time  & GPU & & Time  & GPU  & \\
    \hline
    Endo-FM~\cite{Wang2023FoundationPre-train}            & 6.9  & 221 & &50.4  & 891  & 51.5 \\
    EndoViT~\cite{Batic2024EndoViT:Images}            & 6.7  & 215& & 47.8  & 874  & 45.4 \\
    SurgeNet~\cite{Jaspers2026ScalingModels}           & 12.7 & 157 && 61.7  & 941  & 68.8 \\
    GSViT~\cite{Schmidgall2024GeneralSurgery}              & 22.4 & 56  && 105.8 & 178  & 22.0 \\
    \hline
    LemonFM (CN-B) & 7.8  & 194 && 54.8  & 1068 & 74.7 \\
    LemonFM (CN-L) & 8.9  & 387  && 58.7  & 1782  & 76.4  \\
    \hline
    \end{tabularx}
\end{table}

% \begin{table}[t]
% \centering
% \caption{Computational cost analysis. Inference and training times are measured in milliseconds (ms). Memory usage is reported in Megabytes (MB).}
% \label{tab:compute_cost}
% \resizebox{\linewidth}{!}{% Resize to fit column width if necessary
% \begin{tabular}{lccccc}
% \toprule
% \multirow{2}{*}{\textbf{Model}} & \multicolumn{2}{c}{\textbf{Inference}} & \multicolumn{2}{c}{\textbf{Training}} & \multirow{2}{*}{\textbf{Acc (\%)}} \\
% \cmidrule(lr){2-3} \cmidrule(lr){4-5}
%  & Time (ms) & GPU (MB) & Time (ms) & GPU (MB) & \\
% \midrule
% Endo-FM~\cite{Wang2023FoundationPre-train}           & 6.9  & 221 & 50.4  & 891  & 51.5 \\
% EndoViT~\cite{Batic2024EndoViT:Images}           & 6.7  & 215 & 47.8  & 874  & 45.4 \\
% SurgeNet~\cite{Jaspers2025ScalingModels}          & 12.7 & 157 & 61.7  & 941  & 68.8 \\
% GSViT~\cite{Schmidgall2024GeneralSurgery}            & 22.4 & 56  & 105.8 & 178  & 22.0 \\
% \midrule
% LemonFM (CN-B) & 7.8  & 194 & 54.8  & 1068 & 74.7 \\
% LemonFM (CN-L)& 8.9  & 387 & 58.7  & 1782 & 76.4 \\
% \bottomrule
% \end{tabular}
% }
% \end{table}

\subsection{Computational Cost Comparison}
%
%We analyze the computational efficiency of our method compared to baselines. All measurements were conducted on an NVIDIA A100 GPU using $224\times224$ images in FP16 precision. Metrics include inference time (per image), training time (per step with batch size 8), peak GPU memory usage, and linear probing accuracy on AutoLaparo phase recognition.
%
We analyze the computational efficiency of our method compared to previous state-of-the-art baselines in Table~\ref{tab:compute_cost}. To support diverse computational constraints, we provide LemonFM with two backbone variants: ConvNeXt-Base and ConvNeXt-Large, both of which will be publicly released.

All measurements were conducted on an NVIDIA A100 GPU using $224\times224$ images in FP16 precision. Metrics include inference time (per image), training time (per step with batch size 8), peak GPU memory usage, and linear probing accuracy on AutoLaparo phase recognition.
While strictly lightweight baselines like EndoViT achieve the lowest inference latency (6.7 ms), they suffer from limited representational capacity (45.4\% accuracy). In contrast, our LemonFM (CN-Base) maintains highly competitive efficiency (7.8 ms inference) while delivering a substantial performance leap of nearly 30 percentage points (74.7\%). Even our larger variant (CN-Large) remains efficient (8.9 ms) while further pushing accuracy to 76.4\%. This demonstrates that LemonFM achieves significant performance gains while maintaining practical computational efficiency suitable for real-time applications.

\section{Evaluation of LemonFM-Vid}
To evaluate the performance of our video classifier, LemonFM-Vid, we split the LEMON dataset by videos, with 3369 videos for training and 825 videos for testing. Our vanilla LemonFM was trained on all the LEMON frames; therefore, to evaluate the video classification performance fairly, we trained a new LemonFM model from scratch using only the frames of the 3369 videos in the training set that are accessible to all other classifiers in the comparison.
%As LemonFM was trained with an auxiliary signal containing procedure type information obtained from K-NN process, the pertaining of LemonFM leveraged the whole dataset of LEMON. For a fair comparison with other video classifiers, we trained a new model LemonFM-Vid only on the training and validation sets and then compared them on the test set. 
%Original LemonFM was trained using an auxiliary supervisory signal that incorporates procedure type information by leveraging nearest neighbor from different videos of the same procedure. Consequently, directly employing LemonFM may result in data leakage. To mitigate this risk, we restrict model pretraining exclusively to videos from the training and validation sets of each respective task.

\section{Dataset and Licensing}
We adhere to the same practices as other datasets created from YouTube and various sources on the Web, such as ImageNet~\cite{Deng2009ImageNet:Database}, Kinetics~\cite{Carreira2017QuoDataset}, YouTube-VIS~\cite{Yang2019VideoSegmentation}, YouTube-8M~\cite{Abu-El-Haija2016YouTube-8M:Benchmark}, Insect-1M~\cite{Nguyen2024Insect-Foundation:Understanding}, Moments-in-Time~\cite{Monfort2020MomentsUnderstanding}, Tai-Chi-HD~\cite{Siarohin2019FirstAnimation}, HD-Villa-100M~\cite{Xue2022AdvancingTranscriptions}, and AVSpeech~\cite{Ephrat2018LookingSeparation}.
Specifically, we provide the code to download and generate the LEMON dataset, a list of links to the original YouTube videos, and the corresponding annotations.
Researchers working in academic institutions can request direct access to the LEMON dataset (1 fps version) in LMDB format for non-commercial purposes.
We will also provide an online form for YouTube video authors to opt out of our LEMON dataset.
The LEMON dataset is provided under the Creative Commons Attribution 4.0 International (CC BY 4.0) license.

\FloatBarrier

{
    \small
    \bibliographystyle{ieeenat_fullname}
    \bibliography{main}
}

\end{document}